\documentclass[letterpaper]{article} 
\usepackage{aaai25}  
\usepackage{times}  
\usepackage{helvet}  
\usepackage{courier}  
\usepackage[hyphens]{url}  
\usepackage{graphicx} 
\usepackage{natbib}  
\usepackage{caption} 
\usepackage{algorithm}
\usepackage{algorithmic}
\usepackage{bbding}
\usepackage{array}
\usepackage{amssymb}
\usepackage{romannum}
\usepackage{booktabs}
\usepackage{multirow}
\usepackage{amsmath}

\usepackage{newfloat}
\usepackage{listings}
\DeclareCaptionStyle{ruled}{labelfont=normalfont,labelsep=colon,strut=off} 
\floatstyle{ruled}
\newfloat{listing}{tb}{lst}{}
\floatname{listing}{Listing}
\title{MonoBox:~Tightness-free Box-supervised Polyp Segmentation using Monotonicity Constraint}
\author{
    Qiang Hu\textsuperscript{\rm 1}, Zhenyu Yi\textsuperscript{\rm 2}, Ying Zhou\textsuperscript{\rm 1}, Fan Huang\textsuperscript{\rm 3}, Mei Liu\textsuperscript{\rm 4}, Qiang Li\textsuperscript{\rm 1}, Zhiwei Wang\textsuperscript{\rm 1}\thanks{Corresponding author.}\\
}
\affiliations{
    \textsuperscript{\rm 1}Wuhan National Laboratory for Optoelectronics, Huazhong University of Science
and Technology \\ \textsuperscript{\rm 2} School of Engineering Sciences, Huazhong University of Science and Technology \\ \textsuperscript{\rm 3} Wuhan United Imaging Healthcare Surgical Technology Co., Ltd. \\ \textsuperscript{\rm 4} Tongji Medical College of Huazhong University of Science and Technology\\
    \{huqiang77, zwwang\}@hust.edu.cn
}

\usepackage{bibentry}

\begin{document}

\maketitle

\begin{abstract}
We propose MonoBox, an innovative box-supervised segmentation method constrained by monotonicity to liberate its training from the user-unfriendly box-tightness assumption.
In contrast to conventional box-supervised segmentation, where the box edges must precisely touch the target boundaries, MonoBox leverages imprecisely-annotated boxes to achieve robust pixel-wise segmentation.
The `linchpin' is that, within the noisy zones around box edges, MonoBox discards the traditional misguiding multiple-instance learning loss, and instead optimizes a carefully-designed objective, termed \emph{monotonicity constraint}.
Along directions transitioning from the foreground to background, this new constraint steers responses to adhere to a trend of monotonically decreasing values.
Consequently, the originally unreliable learning within the noisy zones is transformed into a correct and effective monotonicity optimization.
Moreover, an adaptive label correction is introduced, enabling MonoBox to enhance the tightness of box annotations using predicted masks from the previous epoch and dynamically shrink the noisy zones as training progresses.
We verify MonoBox in the box-supervised segmentation task of polyps, where satisfying box-tightness is challenging due to the vague boundaries between the polyp and normal tissues.
Experiments on both public synthetic and in-house real noisy datasets demonstrate that MonoBox 
exceeds other anti-noise state-of-the-arts by improving Dice by at least 5.5\% and 3.3\%, respectively.
Codes are at \url{https://github.com/Huster-Hq/MonoBox}.
\end{abstract}

%

\section{Introduction}
\label{sec:intro}
Colorectal Cancer (CRC) threats to human health worldwide, and colonoscopy is a golden-standard of identifying and resecting early polyps~\cite{haggar2009colorectal,ji2024frontiers}.
Recently, numerous deep learning-based polyp segmentation methods \cite{fan2020pranet,zhao2021automatic,dong2021polyp,zhang2022hsnet,ji2023deep} have been proposed to assist the accurate resection and workload reduction.
However, they are mostly fully-supervised and thus require pixel-level mask labels, which are time-consuming and expensive to acquire.
\begin{figure}[t]
\centering
\includegraphics[width=0.45\textwidth]{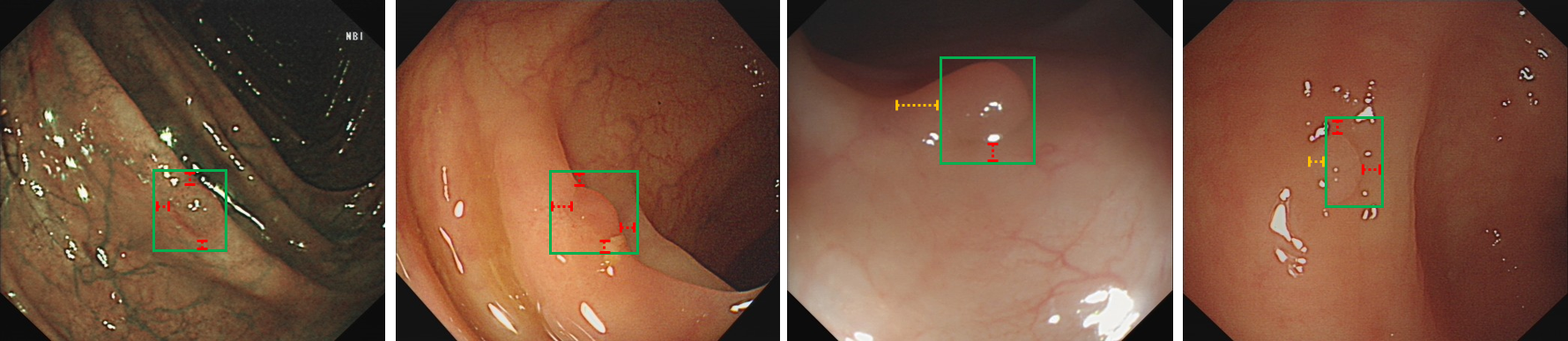}
\caption{Examples of non-tight box annotations produced by endoscopists in the real annotation process. The red dashed lines indicate regions where the annotation is too wide, and the yellow dashed lines indicate regions where the annotation is too narrow.}
\label{fig0}
\end{figure}

To reduce the annotation cost, weakly-supervised segmentation (WSS) methods are studied, aiming to train segmentation models with more cost-effective labels, such as image-level categories \cite{ahn2019weakly}, points \cite{cheng2022pointly}, and bounding boxes \cite{tian2021boxinst,wang2021bounding}.
Among them, box-supervised segmentation (BSS) methods achieve the closest performance to fully-supervised methods, and thus attract dominant research attentions.
The prevailing idea of BSS is called multiple-instance learning (MIL), which views each pixel as an instance, and defines a pixel-width image column or row as a positive bag if it crosses the annotated box, or negative bag otherwise.
By pooling the instances' predictions as the corresponding bag-level prediction, the segmentation model can be trained to produce pixel-level results in the optimization of bag classification.
However, the existing MIL-based BSS methods~\cite{tian2021boxinst,wang2021bounding,cheng2023boxteacher,wei2023weakpolyp,wang2023iboxcla} mostly, if not all, are based on a box-tightness assumption, that is, the box's edges must precisely touch the target to make sure the bag-level labels are accurate.
This severely inhibits the practical application of polyp datasets.
In the polyp context, as shown in Figure~\ref{fig0}, the characteristics of polyps, such as blurred boundaries, low contrast with normal tissues, and small sizes, bring ambiguities to annotators and lead to inaccurate (i.e., non-tight) box annotations.
Therefore, improving the tolerance of BSS methods to inaccurate boxes can alleviate the annotation difficulty and thus is of urgency for the clinical usage of polyp segmentation.
\begin{figure}[t]
\centering
\includegraphics[width=0.45\textwidth]{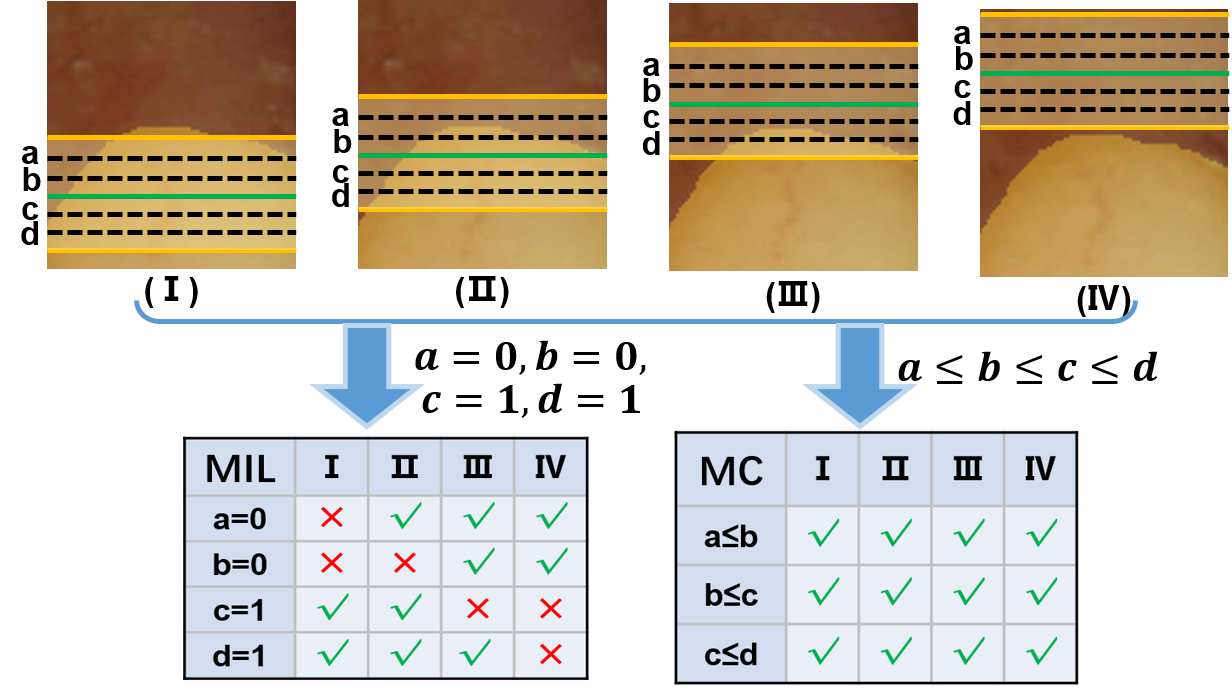}
\caption{In cases (\Romannum{1}, \Romannum{2}, \Romannum{3}, \Romannum{4}) of four typical noisy box annotations (green line), our proposed monotonicity constraint (MC) provides correct constrains for four sampled bags (black dashed line) from the unconfident region (the region between two yellow lines), but the traditional MIL lead to incorrect constraints. For brevity, we only visualize the local region of the upper boundary of the box annotation.}
\label{fig1}
\end{figure}

The above objective can fall into the scope of noise learning, where the noise is induced by box non-tightness.
A straightforward solution is to rectify the boxes to enhance the tightness~\cite{song2019selfie,xu2021training}, but this usually requires a set of clean data as reference, which is beyond the scope of this work. Without clean reference, one promising solution is to apply noise-tolerant classification losses~\cite{wang2019symmetric,ma2020normalized} for addressing the inaccurate bag-level labels in MIL.
In addition to the loss term, sampling more accurate bags is another direction~\cite{wang2022polar,zhu2023contrast}.
However, no matter improving classification losses or sampling strategies in MIL, these solutions ignore the spatial correlation of noises (i.e., incorrect bags), showing weakness in the task of segmentation.

In this paper, we propose MonoBox that constrains monotonicity for tightness-free BSS, which is motivated by two characteristics of the spatial distribution of incorrect bags: (1) the confusing bags are sampled nearby the non-tight box edges, and (2) the probability of sampling a positive bag decrease from inside to outside across the box edges.
Specifically, we first define the regions near the box edges as unconfident regions and the others as confident ones.
For bags sampled from the confident regions, we adopt traditional MIL constraints.
For bags sampled from the unconfident regions, we propose a novel monotonicity constraint (MC), which does not force hard predictions aligning the unreliable labels, but instead encourages a monotonicity trend that the inner response should be higher than the outer response.
Principally, MC finds a soft but reliable surrogate objective when the precise ground-truth is absent, and the objective conforms to the expected spatial distribution pattern.
As evidenced by the four typical noisy cases in unconfident regions (see Figure~\ref{fig1}, for the bags (i.e., a,b,c,d) across the inaccurate box edge, the MIL's classification loss may result in misguiding supervisions, while MC can always derive meaningful gradients for optimization.

Moreover, as the model gradually gains the ability of distinguishing polyp pixels, the MC's imposing areas should dynamically change correspondingly. Thus, we further introduce a label correction strategy to gradually replace the noisy boxes with the predicted masks' bounding boxes, and shrink the unconfident regions, as the training progresses.
This improves the box tightness and increases the learning efficacy of MonoBox.

In summary, our major contributions are as follows.
\begin{itemize}
 	\item[$\bullet$] We propose a tightness-free box-supervised polyp segmentation method, namely MonoBox, which can be effectively trained to precisely segment polyps using noisy box annotations.
	\item[$\bullet$] We propose a new monotonicity constraint (MC) to convert the confusing supervisions into reliable ones on the unconfident regions, and a label correction strategy to dynamically improve the box-tightness during training.
	\item[$\bullet$] We conduct extensive and comprehensive experiments on both public and in-house datasets. The public dataset contains our synthetic controllable noises, and the in-house shows real non-tightness patterns. The comparison results demonstrate that our method can generally improve the robustness of MIL-based BSS methods for non-tight box annotations, and its superiority over the anti-noise state-of-the-arts with an increase of Dice by at least 5.5\% and 3.3\% on the public and in-house dataset, respectively.
\end{itemize}

\section{Related Work}
\subsection{Fully-supervised Polyp Segmentation}
The polyp segmentation methods mostly rely on the fully-supervised learning to train their models.
For example, PraNet \cite{fan2020pranet} generated a global map as the initial guidance region and then used the reverse attention module to refine the segmentation results by using the multilevel features.
SANet \cite{wei2021shallow} adopted a shallow attention module and a color exchange operation to remove background noise and reduce the interference of image color to segmentation respectively.
Polyp-PVT \cite{dong2021polyp} utilized a transformer encoder, a camouflage identification module, and a similarity aggregation module, to effectively suppress noises in the features and significantly improve their expressive capabilities.
However, these methods all require pixel-level polyp annotations, which is difficult and time-consuming to acquire than box-level annotations, impeding the application on large-scale datasets.

\subsection{Weakly-supervised Segmentation using Boxes}
For reducing the cost of annotation, box-supervised segmentation (BSS) is extensively studied.
These methods can be broadly classified into two categories.
One is generating pseudo-labels and the other is based on multiple-instance learning (MIL).

Earlier, most of the methods were based on generating pseudo-labels.
For example, BoxSup~\cite{dai2015boxsup} and Box2Seg~\cite{kulharia2020box2seg} used MCG~\cite{williams2004guidelines} and  GrabCut~\cite{rother2004grabcut}, respectively to generate refined pixel-level masks as pseudo labels based on ground truth (GT) boxes.
Mahani~\emph{et al.}~\cite{mahani2022bounding} proposed to reduce the loss weight of high entropy regions according to the predicted results to reduce error propagation during training. However, these methods rely heavily on the quality of pseudo labels and lack a stable and accurate guidance.

Recently, an idea based on MIL has been adopted by a wide range of methods due to its effectiveness~\cite{hsu2019weakly,tian2021boxinst,lan2023vision,cheng2023boxteacher,wang2023iboxcla}.
It takes the compactness of the box as a priori and considers the maximum value of any row or column of pixels within the box as a positive and the maximum value of any row or column of pixels not passing through the box as a negative.
The implementation details vary from method to method, such as projection into a 1D vector~\cite{tian2021boxinst,lan2023vision,cheng2023boxteacher} or reconstruction into a 2D box-like mask~\cite{wei2023weakpolyp,wang2023iboxcla}.
With MIL loss as the pipeline, some auxiliary constraints, such as color parwise affinity~\cite{tian2021boxinst}, multi-scale consistency~\cite{wei2023weakpolyp} are introduced to further improve the performance.
Among them, IBoxCLA~\cite{wang2023iboxcla} proposed Contrastive Latent-Anchors (CLA), which enhances the feature contrast between the polyp and the surrounding normal tissue, and achieves state-of-the-art (SOTA) in box-supervised polyp segmentation.
However, these methods are extremely dependent on the box tightness assumption, which reduces the cost-effective advantage of BSS methods.

\subsection{Learning with Noisy Labels}
The task that training accurate models using noisy labels has been an active research area.
In the classification realm, various techniques were proposed, such as label correction~\cite{ma2018dimensionality,song2019selfie}, noise-tolerant loss function~\cite{ghosh2017robust,ma2020normalized}, and data cleaning~\cite{han2018co,jiang2018mentornet}.
Most of them can be translated to segmentation task which can be regarded as pixel-level classification.
In the detection realm, He \emph{et al.}~\cite{he2019bounding} proposed KL loss for learning localization variance to alleviate the interference of ambiguity boxes on the detector.
Xu \emph{et al.}~\cite{xu2021training} introduced a meta-learning method to deal with noisy labels by utilizing a few clean samples.
OA-MIL and SSD-Det~\cite{wu2023spatial} proposed to leverage clean class labels as guidance signals for refining inaccurate bounding boxes.
However, there are drawbacks when applying to segmentation.
The solutions for classification do not consider the structure information of the object, while those for detection can not well guarantee the tightness in corrected boxes.

At present, there are only a few studies explicitly aiming at BSS with noisy box labels.
PolarT~\cite{wang2022polar} adopt MIL on the polar transformed image, which reduces the number of incorrect bags but biases the model toward simple instances.
FSRM~\cite{zhu2023contrast} first generated pseudo-masks based on the noisy boxes, and then relied on the pseudo-masks to guide the bag sampling in MIL. However, the noises also can be inherited in the pseudo-masks, negatively impacting the following MIL learning consequently.
In this paper, we aim to seek a noise-tolerant constraint to replace the unreliable MIL-loss on the non-tight regions, and view the constraint as a plug-and-play module to boost the robustness of the current MIL-based SOTAs to tightness-free box annotations.

\section{Method}
Figure~\ref{fig2}(b) gives the overview of MonoBox, which employs the most advanced and efficient image-level MIL fashion~\cite{wei2023weakpolyp,wang2023iboxcla} by optimizing a proxy map rather than individual sampled bags for box-supervised segmentation (BSS), and introduces a new monotonicity constraint (MC) to improve the optimization reliability on the noisy regions. Meanwhile, MonoBox designs a label correction strategy to dynamically improve the tightness of box annotations. In the following, we first recap the optimization of proxy map, and then explain the key components of MonoBox, and at last briefly discuss the extensibility of MC to other box-supervised MIL frameworks of segmentation.

\subsection{Optimizing Proxy Map for BSS}
Given a colonoscopy image $I \in \mathbb{R}^{3 \times H \times W}$ and its GT box annotation $B=(x_{lt},y_{lt},x_{rb},y_{rb})$, where $(x_{lt},y_{lt})$ and $(x_{rb},y_{rb})$ represent the coordinates of the top-left and bottom-right points of box, respectively, the GT box-filled mask $b\in{\{0, 1\}}^{H \times W}$ is created by assigning 1 within $B$ and 0 outside $B$, and a segmented map $m \in (0,1)^{H \times W}$ is obtained by applying the segmentation model on $I$.

\begin{figure*}[t]
\centering
\includegraphics[width=0.83\textwidth]{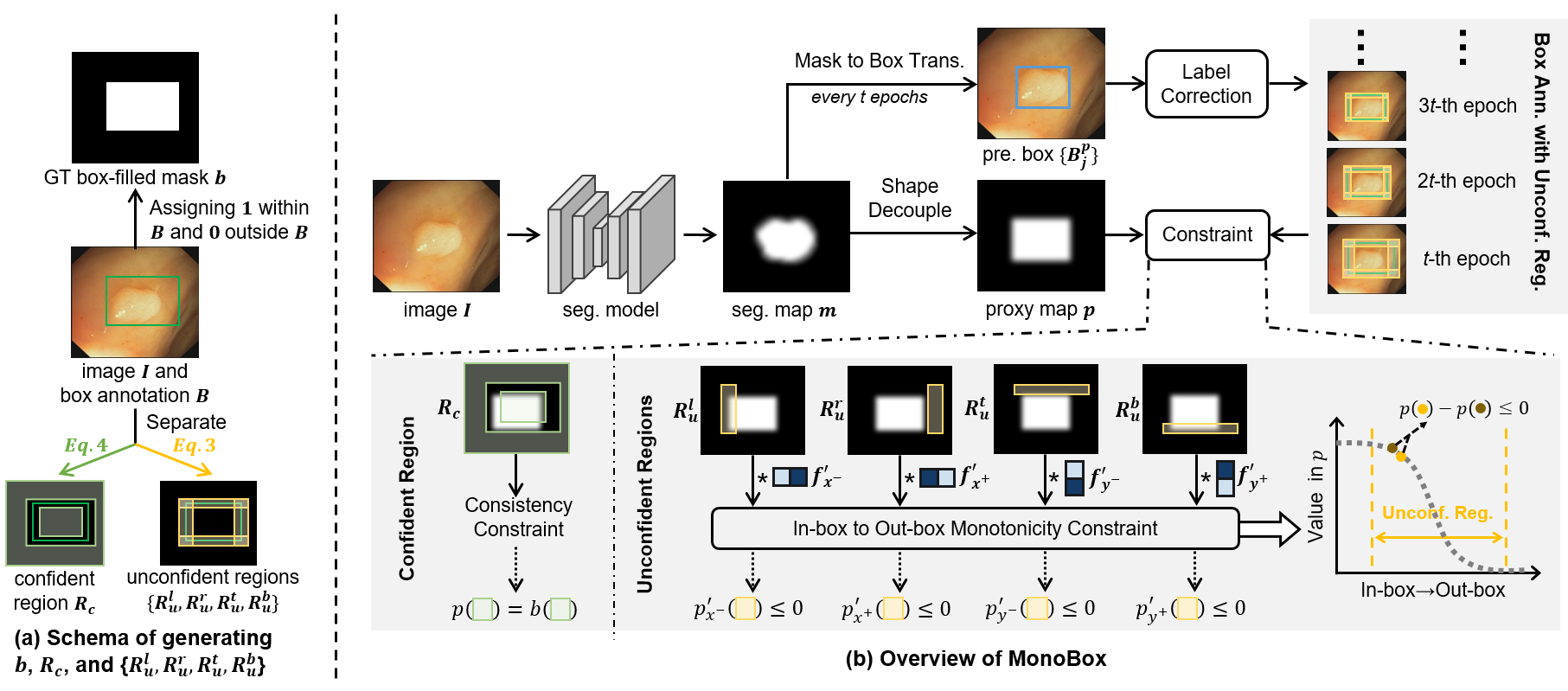}
\caption{(a) Schema of generating GT box-filled mask (assigning 1 within box annotations and 0 outside box annotations), confident region (as Eq.~\ref{eq4}) and unconfident regions (as Eq.~\ref{eq3}). (b) Overview of our proposed MonoBox. For the proxy map decoupled from the segmented map, we first define the confident and unconfident regions, and then adopt the consistency constraint and monotonicity constraint on them, respectively. Moreover, we utilize a strategy of label correction strategy to dynamically improve the tightness of box annotations}
\label{fig2}
\end{figure*}

The GT mask $b$ and the predicted map $m$ have discrepant representation, which impedes a direct optimization between them. To bridge the gap, $m$ should firstly be converted into a proxy map $p$ using the following equation:
\begin{equation}\small
p[i,j] = \mathtt{max}(m[i,:]) \cdot \mathtt{max}(m[:,j]),
\label{eq1}
\end{equation}
where $i$ and $j$ represent the pixel indexes.
The proxy map $p$ decouples the shape information from the segmentation map, and thus can be optimized directly using the GT box-filled mask, as illustrated in Figure~\ref{fig2}(b).
The typical optimization objective is the consistency constraint (e.g., Dice) between $b$ and $p$, which is formulated as follows: 
\begin{equation}\small
    \mathcal{L}_{CC} = -\frac{2\times |b  \cap p|}{|b| + |p|}.
    \label{eq2}
\end{equation}

The consistency constraint relies on the tightness as a prior, which is hard to be satisfied for polyps and thus renders the above consistency constraint unreliable.

\subsection{Separating Confident and Unconfident Regions}
Based on the fact that the noise in the $m$ caused by non-tight box distributes around the edges, we divide the proxy map into confident and unconfident regions.
Specifically, as shown in Figure~\ref{fig2}(a), there are four unconfident regions $\{R_u^l, R_u^r, R_u^t, R_u^b \} \in \mathbb{R}^{1 \times H \times W}$, corresponding to the left, right, top, and bottom edges of the GT box, respectively.
The areas of four unconfident regions are determined by the width and height of the GT box annotation and an unconfident scale $\lambda$.
For example, $R_u^l$ can be calculated as:
\begin{equation}\small
R_{u}^{l}\left( {x,y} \right) = 
\left\{ \begin{array}{lr}
{1,~~x \in \left[ {x_{lt} - \lambda \cdot w,x_{lt} + \lambda \cdot w} \right]} & \\
{~~~~~and~y \in \left[ {y_{lt} - \lambda \cdot h,y_{rb} + \lambda \cdot h} \right],}& \\
{0,~~otherwise.}
\end{array} \right.
\label{eq3}
\end{equation}
where $w=x_{rb}-x_{lt}$ and $h=y_{rb}-y_{lt}$ are the width and height of the GT box, respectively.
The confident region is mutually exclusive with four unconfident regions, which can be formulated as follows:
\begin{equation}\small
R_c=\mathbf{J}-(R_u^l \cup R_u^r \cup R_u^t \cup R_u^b),
\label{eq4}
\end{equation}
where $\mathbf{J}$ is a map of size $H \times W$ with all ones.
On the confident region, we adopt the consistency constraint, which is calculated in Eq.~\ref{eq2}.
On the unconfident region, we use our proposed monotonicity constraint, which is detailed in the next part.

\subsection{Monotonicity Constraint on Unconfident Regions}
\label{3.3}
Monotonicity constraint (MC) encourages a monotonicity trend that the box-inner response should be higher than the box-outer response.
This is inspired by the fact that the closer to the box center, the higher probability of sampling positive bags, which is also satisfied for those non-tight boxes. This fact implies that, in the first-order derivative function of the direction pointing to the box center, the gradient should not exceed zero value.
To this end, we first design four operation kernels, $f_{x^{-}}^{'}$, $f_{x^{+}}^{'}$, $f_{y^{-}}^{'}$, and $f_{y^{+}}^{'}$, which are formulated as:
\begin{equation}\small
    f_{x^{-}}^{'} = \begin{bmatrix}
1  \\
-1 
\end{bmatrix}^{T}
f_{x^{+}}^{'} = \begin{bmatrix}
{- 1}\\
{1}
\end{bmatrix}^{T}
f_{y^{-}}^{'} = \begin{bmatrix}
1 \\
{- 1}
\end{bmatrix}
f_{y^{+}}^{'} = \begin{bmatrix}
{- 1}\\
1
\end{bmatrix},
\label{eq5}
\end{equation}
and then we take {$f_{x^{-}}^{'}$, $f_{x^{+}}^{'}$, $f_{y^{-}}^{'}$, and $f_{y^{+}}^{'}$} to compute the first-order gradient maps on the proxy map $p$ for $R_u^l$, $R_u^r$, $R_u^t$, and $R_u^b$, respectively.
{$p_{x^{-}}^{'}$, $p_{x^{+}}^{'}$, $p_{y^{-}}^{'}$, and $p_{y^{+}}^{'}$} are the four corresponding gradient maps, and utilized to calculate losses by comparing them with zeros for monotonicity constraint (MC loss).
Taking the MC loss in $R_u^l$ as an example, the formula is as follows:
\begin{equation}\small
    \mathcal{L}_{MC}^{l}= {\sum\limits_{i,j|R_{u}^{l}{[i,j]} = 1}{max\left( p_{x^{-}}^{'}[i,j],0 \right).}}
    \label{eq6}
\end{equation}
Similarly, we calculate the MC loss for the other three unconfident regions, denoted as $\mathcal{L}_{MC}^{r}$, $\mathcal{L}_{MC}^{t}$, and $\mathcal{L}_{MC}^{b}$.
Therefore, the complete MC loss is formulated as:
\begin{equation}\small
\mathcal{L}_{MC} = \mathcal{L}_{MC}^{l} + \mathcal{L}_{MC}^{r} + \mathcal{L}_{MC}^{t} + \mathcal{L}_{MC}^{b}.
\label{eq7}
\end{equation}

\subsection{Label Correction for Dynamic Training}
\label{3.4}
We introduce label correction to maximize the learning efficacy.
This is motivated by the observation that the segmentation model guided by the above MC loss can gradually gain the ability to classify polyp pixels, yielding reasonable segmentation results closer to the true clean box.
Specifically, we transform the segmentation results into boxes $\{ B_j^p, j=0,1,...,M\}$ by finding connected regions and calculating the tightest box
for each region.
Next, we match the predicted boxes $\{ B_j^p\}$ with the GT boxes $\{ B_i\}$.
The rule is that a GT box matches with only one predicted box that has the largest IoU with it, and the IoU must exceed a threshold $\tau$.
We merge the matched GT and predicted boxes to get the corrected boxes, and left the unmatched GT boxes as they are.
The label correction increases the tightness of most GT boxes, and thus lowers noise degree.
Therefore, the unconfident scale $\lambda$ in Eq.~\ref{eq3} should be decreased accordingly.
During the training, we evoke the label correction every $t$ epochs, and dynamically adjust $\lambda$ after each label correction by half reduction.

\subsection{Generality for MIL-based BSS}
\label{gernerality}
Although the implementation of the MC is based on proxy map optimization, we would like to remark that it is easy to expand to other MIL-based variants~\cite{hsu2019weakly,tian2021boxinst} of BSS.
Specifically, for the bags sampled from the unconfident regions, we only need to remember the row or column index of each bag, and the monotonicity constraint is a contrastive constraint of bag pairs with adjacent indexes, that is, the prediction of bag with inside index should be higher than that with outside index. Such process is just simplified by use of the proxy map and the four derivative kernels in our implementation of MonoBox.

\section{Experiments}
\subsection{Implementation Details}
MonoBox is implemented using PyTorch~\cite{paszke2019pytorch} and trained using a single NVIDIA GeForce RTX 3090 GPU with 24GB memory.
We use AdamW~\cite{loshchilov2017decoupled} as the optimizer, and set both the learning rate and weight decay to 0.0001.
We resize the input image into 352 × 352 and set the batch size to 16.
We train the models for 50 epochs in total and evoke the label correction every 10 epochs, i.e., $t=10$.
The unconfident scale $\lambda$ in Eq.~\ref{eq3} and the IoU threshold $\tau$ in label correction are set to 0.2 and 0.7, respectively.

\subsection{Datasets and Evaluation Metrics}
\subsubsection{Public Synthetic Noisy Dataset.}
We select five public polyp datasets used in previous work~\cite{wang2023iboxcla}: ClinicDB \cite{bernal2015wm}, Kvasir-SEG \cite{jha2020kvasir}, ColonDB \cite{tajbakhsh2015automated}, EndoScene \cite{vazquez2017benchmark}, and ETIS \cite{silva2014toward}.
Following~\cite{wang2023iboxcla}, we set 550 samples in ClinicDB and 900 samples in Kvasir as the train-set, and the remaining samples from these two datasets and all samples from the other three datasets as the test-set.
Note that the original annotations for these datasets are ground truth (GT) masks, and we convert the GT masks to boxes by finding the tightest bounding boxes of the connected components.

We consider the boxes converted above to the tight and clean.
For simulating tightness-free boxes, we perturb box coordinates of these clean boxes. Specifically, let $(x_c, y_c, w, h)$ denote the center $x$ coordinate, center $y$ coordinate, width, and height of an clean box.
We simulate an noisy bounding box $(\hat{x_c}, \hat{y_c}, \hat{w}, \hat{h})$ by randomly shifting and scaling the box as follows:
\begin{equation}
\left\{ \begin{matrix}
{\hat{x_c} = x_c + \bigtriangleup{x} \cdot w,~~~\hat{y_c} = y_c + \bigtriangleup{y} \cdot h,} \\
{\hat{w} = \left( {1 + \bigtriangleup{w}} \right) \cdot w,~~~\hat{h} = \left( {1 + \bigtriangleup{h}} \right) \cdot h,}
\end{matrix} \right.
    \label{eq8}
\end{equation}
where $\bigtriangleup{x}$, $\bigtriangleup{y}$, $\bigtriangleup{w}$, and $\bigtriangleup{h}$ follow the normal distribution $\mathcal{N}(0, \sigma^2)$, $\sigma$ is the noise level.

\subsubsection{In-house Real Noisy Dataset.}
The in-house dataset cosists 18,656 colonoscopy images of polyp.
The dataset is from a local hospital, which is split to the train-set (17,350 images) and test-set (1,306 images).
The train-set is provided with box annotations (Figure~\ref{fig0} shows four box-annotated examples), and the test-set is provided with pixel-level annotations by two experts.
Written informed consent was not required for this study as documented clinical colonoscopic images were collected retrospectively and appropriately by anonymizing and deidentifying.
Therefore, the study design was exempted from full review by the Institutional Review Board.

\subsubsection{Evaluation Metrics.}
We first use a threshold of 0.5 to binarize the segmented map of models to obtain the binary mask, and then we employ three widely-used evaluation metrics, including Dice, IoU, and Hausdorff distance (HD) to evaluate the similartity between the segmented and the ground truth (GT) masks.
Among these metrics, Dice and IoU are similarity measures at the regional level, which mainly focus on the internal consistency of segmented objects.
HD can better evaluate the segmentation results at the boundaries.

\subsection{Comparison with Anti-noise SOTAs}
\label{4.2}
\begin{table}[t]\small
\centering
\caption{Comparison segmentation results between MonoBox and other anti-noise methods. The best performance is marked in bold. `LB' means directly training the backbone with noisy datasets, `UB' means training the backbone with clean datasets.} 
{
\scalebox{0.7}{
\begin{tabular}{llcccccc}
	\toprule
	 \multirow{2}{*}{Backbones}& \multirow{2}{*}{Methods} & \multicolumn{3}{c}{Real} & \multicolumn{3}{c}{Synthetic}\\
  &&Dice &IoU &HD(px) &Dice &IoU &HD(px)\\
        \midrule
                \multirow{7}{*}{WeakPolyp}
                &UB& n/a& n/a&  n/a&0.774 &0.694&3.317\\
        &LB&0.775&0.671&3.164&0.700&0.597&4.146\\
        &SSD-Det&0.770&0.651&3.321&0.710&0.599&4.107\\
        &NCE+RCE&0.783&0.684&3.224&0.721&0.615&3.989\\
        &PolarT&0.766&0.654&3.328&0.701&0.591&4.288\\        &FSRM&0.785&0.689&3.174&0.722&0.613&3.853\\
        &MonoBox(Ours)&\textbf{0.804}&\textbf{0.714}&\textbf{2.916}&\textbf{0.763}&\textbf{0.660}&\textbf{3.411}\\
        \midrule
                \multirow{7}{*}{IBoxCLA}
        &UB& n/a& n/a&  n/a &0.827 &0.750 &3.048\\
        &LB&0.803&0.716&2.412 &0.735& 0.628&4.310\\
        &SSD-Det& 0.802& 0.713& 2.324& 0.743& 0.634& 3.898\\
        &NCE+RCE& 0.811& 0.721& 2.203& 0.748& 0.641& 3.737\\
        &PolarT&0.797&0.711&2.538&0.734&0.622&3.664\\
        &FSRM&0.816&0.720&2.242&0.748&0.642&3.664\\
        & MonoBox(Ours)&\textbf{0.849}&\textbf{0.764}&\textbf{1.920}&\textbf{0.803}&\textbf{0.714}&\textbf{3.338}\\
	\bottomrule
\end{tabular}
}
}
\label{table1}
\end{table}
\begin{figure}[t]
\centering
\includegraphics[width=0.47\textwidth]{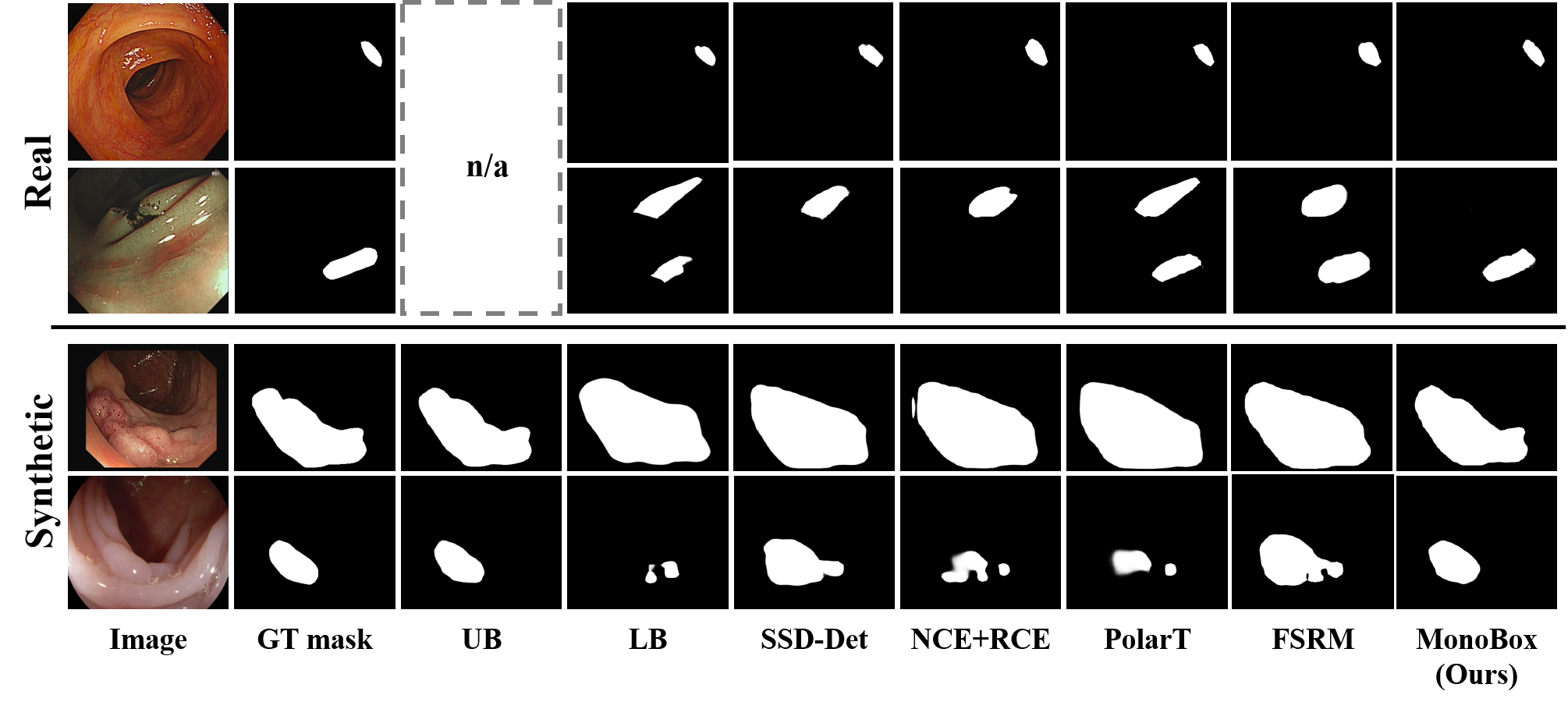}
\caption{Visualization results of the different anti-noise methods using IBoxCLA as the model on the Real noisy dataset and the Synthetic noisy dataset.}
\label{fig4}
\end{figure}
We view MonoBox as a plug-and-play anti-noise method, and compare it with four state-of-the-art (SOTA) anti-noise methods, i.e., SSD-Det~\cite{wu2023spatial}, NCE+RCE~\cite{ma2020normalized}, PolarT~\cite{wang2022polar}, and FSRM~\cite{zhu2023contrast} on the Real noisy dataset and the Synthetic noisy dataset with noise level $\sigma=0.2$.
We train these methods using two box-supervised (BSS) segmentation backbones, i.e., WeakPolyp~\cite{wei2023weakpolyp} and IBoxCLA~\cite{wang2023iboxcla}.
Note that, we denote UB and LB as backbones trained using the clean and noisy datasets, respectively, and since there are no clean annotations in the Real noisy dataset, we do not report results of UB for the Real noisy dataset.

The comparison results are provided in Table~\ref{table1}.
Comparing the results of LB and UB of the two backbones, we can see that on Synthetic noisy dataset, the noise leads a decrease of Dice by 7.4\% and 9.2\% for WeakPolyp and IBoxCLA, respectively.
This exhibits the high sensitivity of the existing BSS methods to noisy boxes.
When using MonoBox for optimization, the performance margin compared to the UB is significantly narrowed for both backbones.
Specifically, Dice of IBoxCLA is improved by 4.6\% and 6.8\% , and Dice of WeakPolyp is improved by 2.9\% and 6.3\%, on the Real and Synthetic noisy datasets, respectively.

Moreover, our method achieves the best performance among all anti-noise methods.
IBoxCLA with our method exceeds that with the second-best method, i.e., FSRM, by 3.3\% and 5.5\% in Dice on the Real and Synthetic noisy datasets, respectively.
WeakPolyp with our method outperforms that with FSRM by 1.9\% and 4.1\% in Dice on the Real and Synthetic noisy datasets, respectively.
This is because that FSRM aims to correct the original noisy boxes using the model's predictions, but the predictions guided by traditional MIL are unreliable under the supervision of noisy labels.
In contrast, the monotonicity constraint of MonoBox can provide reliable supervision even under noisy labels.

Figure~\ref{fig4} shows the visualization results of different methods using IBoxCLA as the backbone.
It can be seen that the model trained with our method can accurately find the inconspicuous polyps and segment more reasonable boundaries.

\subsection{Ablation Study}
\subsubsection{Effectiveness of the key components.}
\begin{table}[t]
\centering
\caption{Ablation study on the effectiveness of the two proposed components, i.e., Monotonicity Constraint (MC) and Label Correction (LC). }
{
\scalebox{0.7}{
\begin{tabular}{m{1.5cm}m{1.5cm}cccccc}
	\toprule
	 \multicolumn{2}{c}{Components} & \multicolumn{3}{c}{Real} & \multicolumn{3}{c}{Synthetic}\\
  MC &LC &Dice &IoU &HD(px) &Dice &IoU &HD(px)\\
	\midrule
	\XSolidBrush& \XSolidBrush& 0.803& 0.716&  2.412& 0.735& 0.628& 4.310\\
        \Checkmark& \XSolidBrush& 0.838& 0.758& 2.053& 0.787& 0.700& 3.655\\
        \XSolidBrush& \Checkmark& 0.809& 0.720& 2.447& 0.744& 0.628& 4.002\\
        \Checkmark& \Checkmark& \textbf{0.849}& \textbf{0.764}& \textbf{1.920}& \textbf{0.803}& \textbf{0.714}& \textbf{3.338}\\
	\bottomrule
\end{tabular}
}
}
\label{table3}
\end{table}
\begin{figure}[t]
\centering
\includegraphics[width=0.42\textwidth]{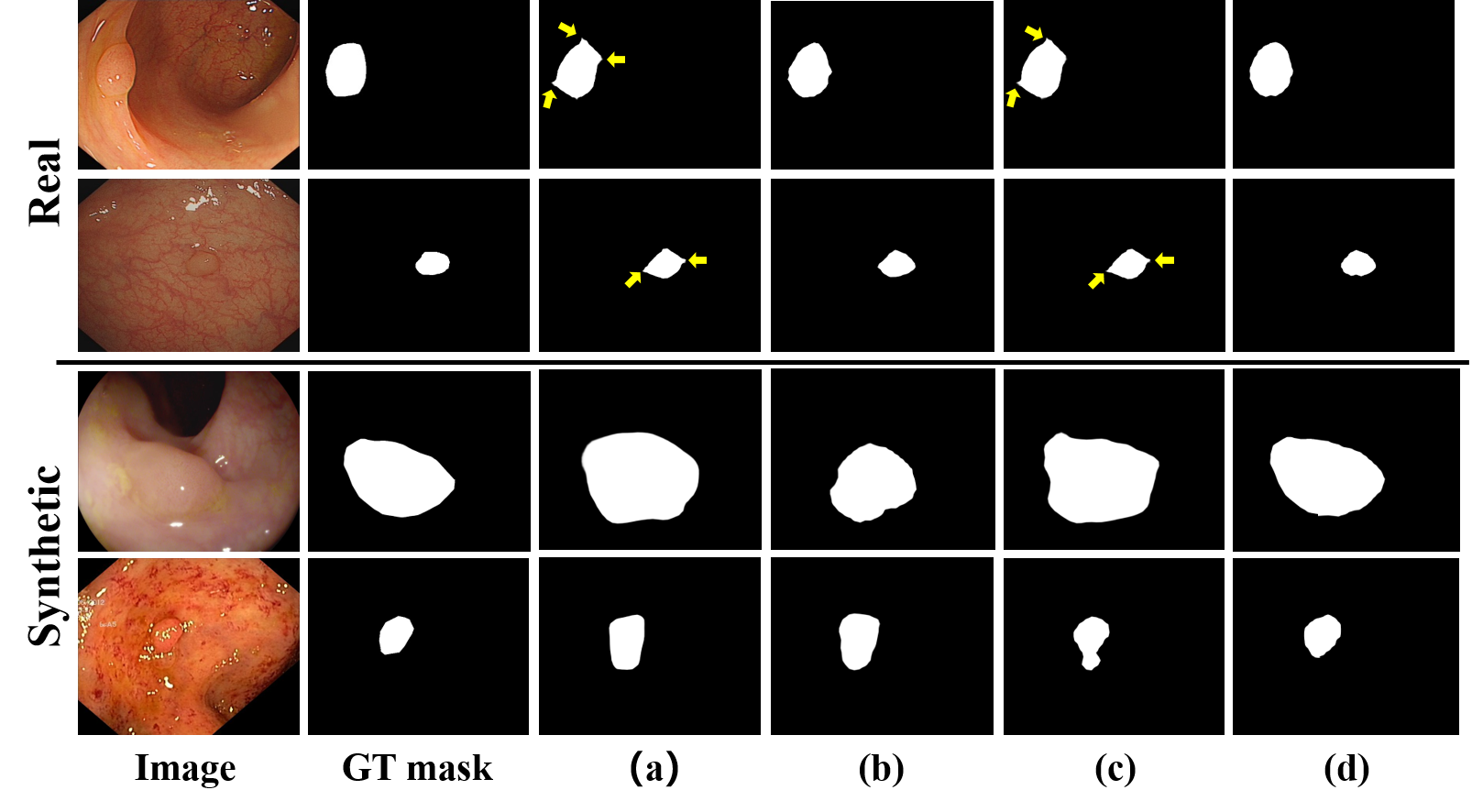}
\caption{Visualization results of the baseline and three variants. (a)-(d) correspond to the 1st to the 4th rows in Table~\ref{table3}, respectively. Yellow arrows show the predicted small thorn-like incorrect regions.}
\label{fig5}
\end{figure}
To verify the effectiveness two key components of MonoBox, i.e., monotonicity constraint (MC) and label correction (LC), we train three variants of MonoBox by disabling MC and/or LC, and IBoxCLA is used as the BSS backbone.
The segmentation results are presented in Table~\ref{table3} and Figure~\ref{fig5}.
Based on these results, two key conclusions can be made as follows:

(1)~By comparing the first two rows in Table~\ref{table3} corresponding to (a) and (b) in Figure~\ref{fig5}, we can see that MC significantly improves the tolerance of the model to noisy box annotations. 
From Figure~\ref{fig5}(a), we can see that on the Real noisy dataset, the baseline tends to produce small thorn-like regions (indicated by yellow arrows) at the boundary.
This is because the doctors often annotated over-sized boxes, and thus the model has to output splinters to touch the incorrect boundaries.
This phenomenon does not occur in the Synthetic noisy dataset, because we use gaussian distributed noise with zero-centered mean value, as shown in Eq.~\ref{eq8}.
It's worth noting that, as shown in Figure~\ref{fig5}(b), MC can mitigate the misdirection of the incorrect box annotations and eliminate erroneous thorn-like regions.

\begin{figure}[t]
\centering
\includegraphics[width=0.33\textwidth]{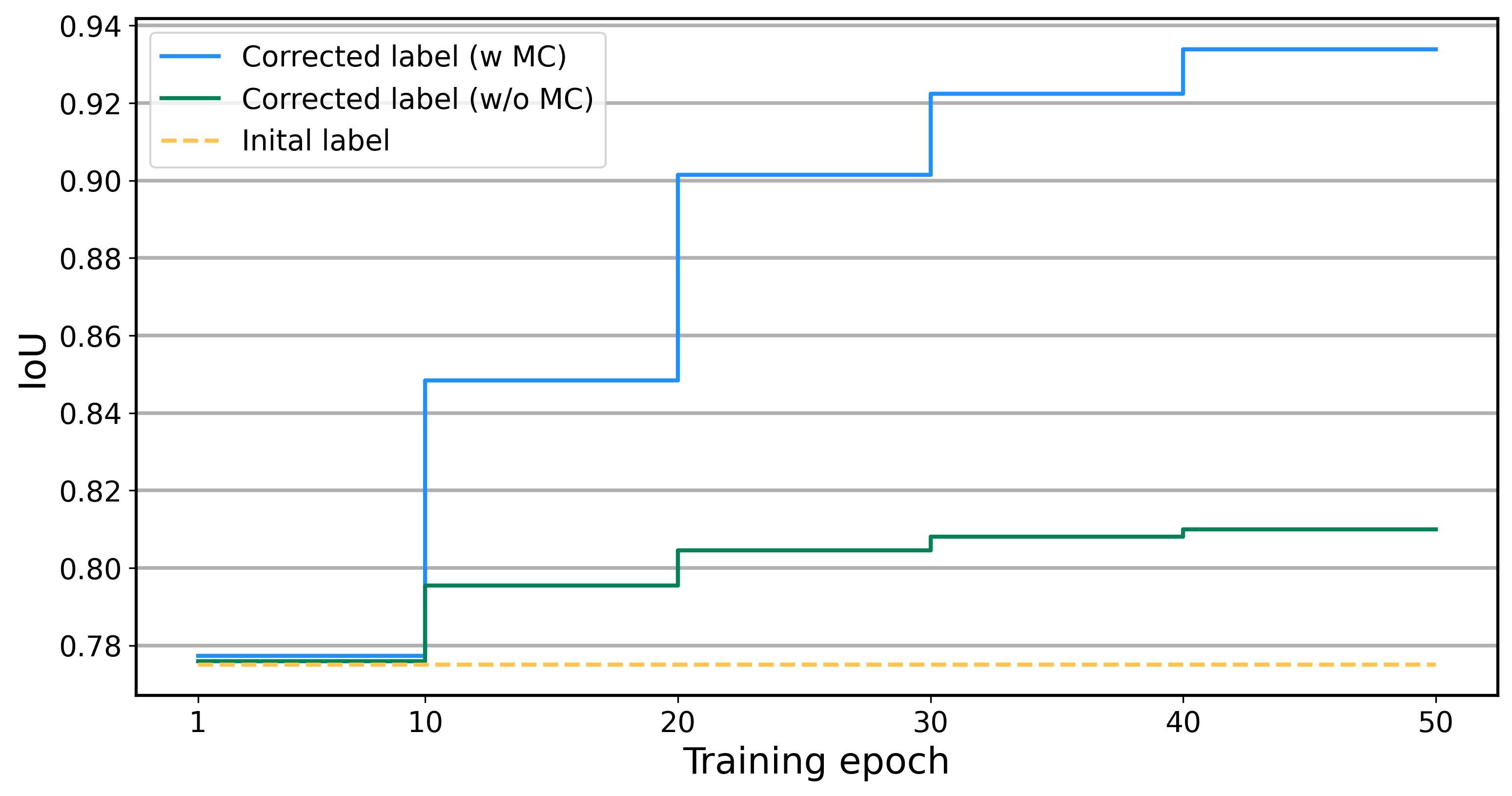}
\caption{Label accuracy curve across the training process on the Synthetic noisy dataset. The ordinate represents the IoU between the training box label and the clean box label. Following the implementation details, the label correction is performed every 10 epochs.}
\label{fig7}
\end{figure}

(2)~By comparing the first and third rows in Table~\ref{table3} and seeing Figure~\ref{fig5}(c), we find that using LC alone brings only limited improvement and fails to address erroneous thorn-like regions.
However, when LC is combined with MC, it can lead more improvement and eliminate the thorn-like regions, as shown in Figure~\ref{fig5}(d).
To clearly show the mechanism of LC, we visualize the label accuracy curve with/without MC across the training process on the Synthetic noisy dataset in Figure~\ref{fig7}.
Without MC, LC hardly improves the accuracy of labels, while with MC, LC significantly and continuously improves the label accuracy.
This is because MC optimizes the constraint in unconfident regions, having the predicted masks more precisely, which can guide LC to improve the tightness of boxes, which in turn can guide more accurate predictions. Therefore, MC and LC complement each other and the combination of them achieves the best performance.

\subsubsection{Different Strategies for Unconfident Region.}
MonoBox essentially defines the unconfident region and improves the reliability of constraint in this region through monotonicity constraint (MC).
Likewise, there are also some strategies that address the constraint of the unconfident region, such as excluding this region in loss calculation, or computing the loss using soft labels.
To further verify the effectiveness of MC, we conduct an experiment to compare the performance of MC with two strategies, denoted as Exclusion and Soft Label, respectively.
We choose 2D-Gaussian Label~\cite{qadir2021toward} as the implementation of Soft Label due to its wide usage.
Note that, to avoid the influence of other components, we disable LC in this experiment.
The quantitative comparison results are presented in Table~\ref{table4}.
As can be seen, using Exclusion hardly improves the performance of the baseline.
This is because simply excluding the unconfident regions could waste the possibly carried valuable information.
Using Soft Label can improve the performance compared to LB, but there is still a large gap compared to UB.
In contrast, our MC significantly improves the baseline's noise tolerance, and achieves better performance than both strategies by remarkable margins.

\begin{table}[t]
\centering
\caption{Comparison results between the Monotonicity Constraint (MC) and two commonly used strategies for addressing the unconfident regions.}
{
\scalebox{0.65}{
\begin{tabular}{llcccccc}
	\toprule
	 \multirow{2}{*}{Methods} & \multicolumn{3}{c}{Real} & \multicolumn{3}{c}{Synthetic}\\
  &Dice &IoU &HD(px) &Dice &IoU &HD(px)\\
	\midrule
        UB&n/a &n/a &n/a &0.827 &0.750 &3.048\\
	LB&0.803 &0.716 &2.412 &0.735 &0.628 &4.310\\
        Exclusion& 0.809& 0.720& 2.371& 0.731& 0.620& 3.940\\
        Soft Label~\cite{qadir2021toward}& 0.816& 0.730& 2.222& 0.749& 0.650& 3.960\\
        MC (Ours)&\textbf{0.838} &\textbf{0.758} &\textbf{2.053} &\textbf{0.787} &\textbf{0.700} &\textbf{3.655}\\
	\bottomrule
\end{tabular}
}
}
\label{table4}
\end{table}

\subsubsection{Hyperparameters Choices.}
Table~\ref{table4} compares our results on the Real noisy dataset with different choices of hyperparameters: confindent scale $\lambda$ in Eq.~\ref{eq3}, IoU threshold $\tau$ and interval epoch $t$ in label correction.
As can be seen, MonoBox using $\lambda=0.2$, $\tau=0.7$, and $t=10$ shows the best performance, respectively.
Note that, MonoBox does not rely on hyperparameter tuning when applied to different datasets.
As shown in Figure~\ref{fig6}, we keep hyperparameter choices constant ($\lambda=0.2$, $\tau=0.7$, $t=10$) on datasets with different noise levels, MonoBox can maintain stable performance and outperform other SOTA methods.
\begin{table}[t]
\centering
\caption{Ablation on $\lambda$, $\tau$ and $t$ on the Real noisy dataset}
{
\scalebox{0.7}{
\begin{tabular}{p{0.7cm}cc|p{0.7cm}cc|p{0.7cm}cc}
	\toprule
  $\lambda$&Dice &HD(px)& $\tau$ &Dice &HD(px)& $t$ &Dice &HD(px)\\
	\midrule
	0.1&0.836&2.108&0.5&0.842&1.994&2&0.840&2.057\\
        0.2& \textbf{0.849}& \textbf{1.920}& 0.6&0.846&1.953&5&0.844&1.987\\
        0.3& 0.844& 1.964& 0.7&\textbf{0.849}&\textbf{1.920}&10&\textbf{0.849}&\textbf{1.920}\\
        0.4&0.838& 2.011& 0.8& 0.845& 1.988&20&0.842&2.043\\
	\bottomrule
\end{tabular}
}
}
\label{table5}
\end{table}

\subsection{Performance under Different Levels of Noise}
\begin{figure}[t]
\centering
\includegraphics[width=0.33\textwidth]{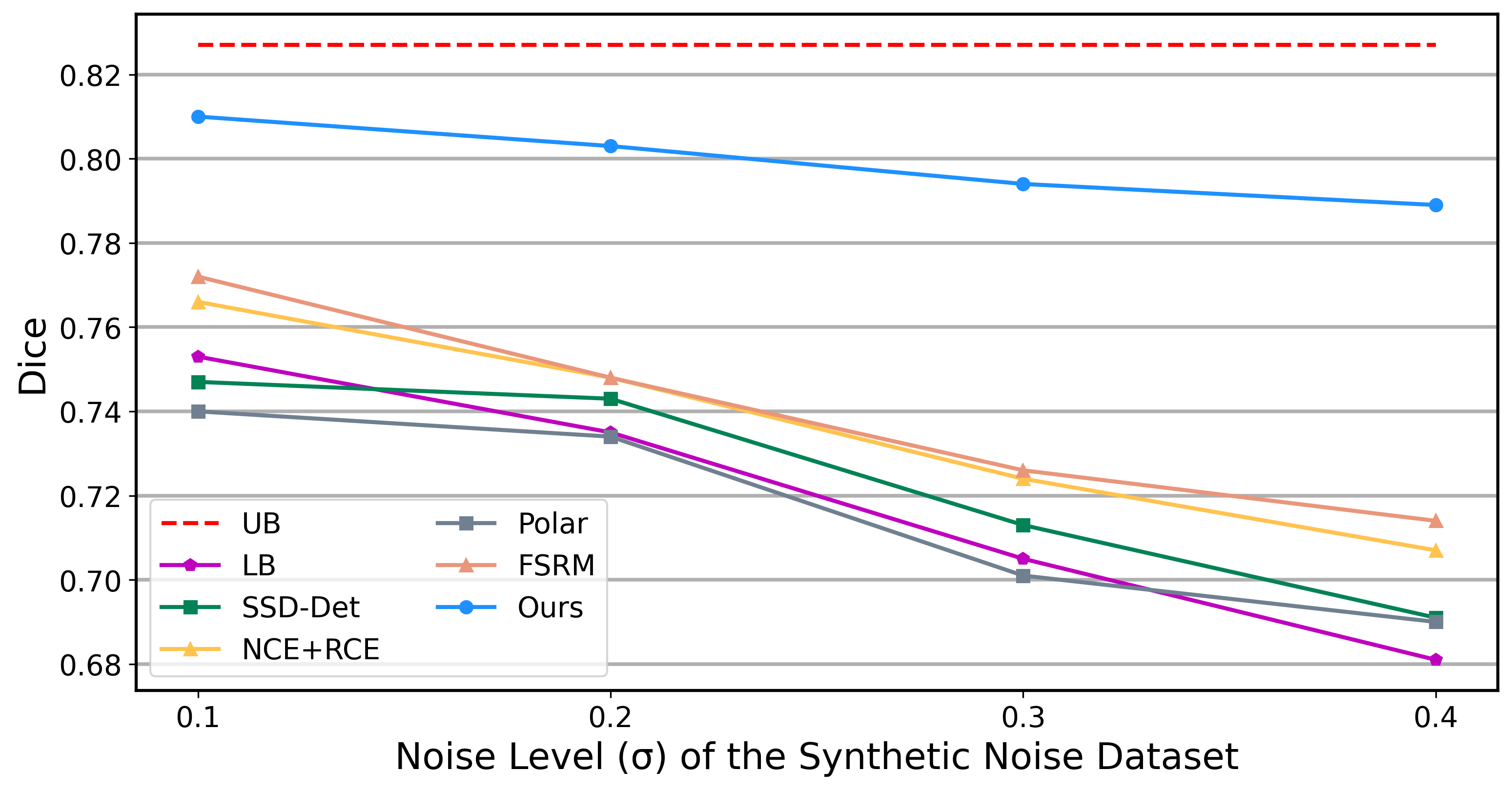}
\caption{Dice of different methods under the different levels of noise on the Synthetic noisy dataset.}
\label{fig6}
\end{figure}
We conduct an experiment to analyze the stability of MonoBox and the previous anit-noise SOTAs against different level of synthetic box noises.
Specifically,  taking IBoxCLA as the BSS backbone, we train the methods multiple times using four Synthetic noisy datasets with different noise levels $\sigma$.
Figure \ref{fig6} illustrates the performance trends in terms of Dice of these methods.
As can be seen, when adding a slight noise $(\sigma=0.1)$, the segmentation backbone suffers from a significant decrease in Dice from 0.827 to 0.753 (UB vs. LB).
Compared to LB, NCE+RCE~\cite{ma2020normalized} and FSRM~\cite{zhu2023contrast} show limited performance gains, and SSD-Det~\cite{wu2023spatial} and PolarT~\cite{wang2022polar} even degrade the performance.
In contrast, MonoBox greatly alleviates the interference of noise, improving Dice by 5.7\% (0.810 vs. 0.753) compared to LB, and almost competitive to UB (0.810 vs. 0.827).
As the noise level increasing, all methods show varying degrees of performance degradation.
However, the curve of MonoBox is stabler and flatter compared to those of other methods, which shows that our method is more robust to noisy boxes.

\subsection{Generality for other scenario and method}
MonoBox is not only applicable to any scenario, but also can be easily implemented with other MIL-based BSS methods.
To verify this, we conduct an experiment on COCO~\cite{lin2014microsoft}, set BoxInst~\cite{tian2021boxinst} as the backbone, and expand the MonoBox to adapt its MIL-based loss, i.e., projection loss.
The tightness-free box annotations are generated by  Eq. ~\ref{eq8}.
As shown the results in Table~\ref{table6}, MonoBox significantly narrows the performance gap between LB and UB and outperforms the second best anti-noise method, i.e, FSRM, by a large margin, which verifies that generality of MonoBox on natural scene and other MIL-based box-supervised segmentation methods.
\begin{table}[t]\small
\centering
\caption{Comparison results on the COCO test-dev split.} 
{
\scalebox{0.7}{
\begin{tabular}{p{1.3cm}p{2.5cm}cccccc}
	\toprule
	 {Backbone}& {Methods} & {AP} & {AP$_{50}$} & {AP$_{75}$} & {AP$_{S}$} & {AP$_{M}$} & {AP$_{L}$}\\
        \midrule
                \multirow{4}{*}{BoxInst}
        &UB& 0.321 & 0.551& 0.324 &0.156 &0.343 &0.435\\
        &LB& 0.248 & 0.480 & 0.225 &0.117 &0.282 &0.329\\
        &FSRM& 0.259 & 0.488& 0.240 & 0.124 & 0.293 &0.344\\
        & MonoBox(Ours)&\textbf{0.308}&\textbf{0.532}&\textbf{0.304} &\textbf{0.149} &\textbf{0.331} &\textbf{0.417}\\
	\bottomrule
\end{tabular}
}
}
\label{table6}
\end{table}

\section{Conclusion and  Future Work}
In this paper, we propose MonoBox, which addresses the limitations of the existing BSS method on tightness-free box annotations.
Inspired by the specific spatial distribution of the noise, we propose MC to provide more reliable optimizations in model training.
Moreover, we propose LC to improve the tightness of the box annotations and dynamically optimize the training process.
Our approach is general and can easily cooperate with modern MIL-based BSS methods.
The comprehensive experiments on the public synthetic noisy dataset and the in-house real noisy dataset demonstrate that MonoBox can effectively solve the realistic problems in clinical practice and has great application value.
Nevertheless, MonoBox currently adopts a uniform unconfident scale for all samples during training, we will explore methods to adaptively perceive the appropriate unconfident scale of each training sample in the future.

\section{Acknowledgments}
This work was supported in part by National Key R\&D  Program of China (Grant No. 2023YFC2414900), Key R\&D Program of Hubei Province of China (No.2023BCB003), Fundamental Research Funds for the Central Universities (2021XX-JS033), Wuhan United Imaging Healthcare Surgical Technology Co., Ltd.

\bibliography{aaai25}

\begin{thebibliography}{43}
\providecommand{\natexlab}[1]{#1}

\bibitem[{Ahn, Cho, and Kwak(2019)}]{ahn2019weakly}
Ahn, J.; Cho, S.; and Kwak, S. 2019.
\newblock Weakly supervised learning of instance segmentation with inter-pixel relations.
\newblock In \emph{Proceedings of the IEEE/CVF conference on computer vision and pattern recognition}, 2209--2218.

\bibitem[{Bernal et~al.(2015)Bernal, S{\'a}nchez, Fern{\'a}ndez-Esparrach, Gil, Rodr{\'\i}guez, and Vilari{\~n}o}]{bernal2015wm}
Bernal, J.; S{\'a}nchez, F.~J.; Fern{\'a}ndez-Esparrach, G.; Gil, D.; Rodr{\'\i}guez, C.; and Vilari{\~n}o, F. 2015.
\newblock WM-DOVA maps for accurate polyp highlighting in colonoscopy: Validation vs. saliency maps from physicians.
\newblock \emph{Computerized medical imaging and graphics}, 43: 99--111.

\bibitem[{Cheng, Parkhi, and Kirillov(2022)}]{cheng2022pointly}
Cheng, B.; Parkhi, O.; and Kirillov, A. 2022.
\newblock Pointly-supervised instance segmentation.
\newblock In \emph{Proceedings of the IEEE/CVF Conference on Computer Vision and Pattern Recognition}, 2617--2626.

\bibitem[{Cheng et~al.(2023)Cheng, Wang, Chen, Zhang, and Liu}]{cheng2023boxteacher}
Cheng, T.; Wang, X.; Chen, S.; Zhang, Q.; and Liu, W. 2023.
\newblock Boxteacher: Exploring high-quality pseudo labels for weakly supervised instance segmentation.
\newblock In \emph{Proceedings of the IEEE/CVF Conference on Computer Vision and Pattern Recognition}, 3145--3154.

\bibitem[{Dai, He, and Sun(2015)}]{dai2015boxsup}
Dai, J.; He, K.; and Sun, J. 2015.
\newblock Boxsup: Exploiting bounding boxes to supervise convolutional networks for semantic segmentation.
\newblock In \emph{Proceedings of the IEEE international conference on computer vision}, 1635--1643.

\bibitem[{Dong et~al.(2021)Dong, Wang, Fan, Li, Fu, and Shao}]{dong2021polyp}
Dong, B.; Wang, W.; Fan, D.-P.; Li, J.; Fu, H.; and Shao, L. 2021.
\newblock Polyp-pvt: Polyp segmentation with pyramid vision transformers.
\newblock \emph{arXiv preprint arXiv:2108.06932}.

\bibitem[{Fan et~al.(2020)Fan, Ji, Zhou, Chen, Fu, Shen, and Shao}]{fan2020pranet}
Fan, D.-P.; Ji, G.-P.; Zhou, T.; Chen, G.; Fu, H.; Shen, J.; and Shao, L. 2020.
\newblock Pranet: Parallel reverse attention network for polyp segmentation.
\newblock In \emph{International conference on medical image computing and computer-assisted intervention}, 263--273. Springer.

\bibitem[{Ghosh, Kumar, and Sastry(2017)}]{ghosh2017robust}
Ghosh, A.; Kumar, H.; and Sastry, P.~S. 2017.
\newblock Robust loss functions under label noise for deep neural networks.
\newblock In \emph{Proceedings of the AAAI conference on artificial intelligence}, volume~31.

\bibitem[{Haggar and Boushey(2009)}]{haggar2009colorectal}
Haggar, F.~A.; and Boushey, R.~P. 2009.
\newblock Colorectal cancer epidemiology: incidence, mortality, survival, and risk factors.
\newblock \emph{Clinics in colon and rectal surgery}, 22(04): 191--197.

\bibitem[{Han et~al.(2018)Han, Yao, Yu, Niu, Xu, Hu, Tsang, and Sugiyama}]{han2018co}
Han, B.; Yao, Q.; Yu, X.; Niu, G.; Xu, M.; Hu, W.; Tsang, I.; and Sugiyama, M. 2018.
\newblock Co-teaching: Robust training of deep neural networks with extremely noisy labels.
\newblock \emph{Advances in neural information processing systems}, 31.

\bibitem[{He et~al.(2019)He, Zhu, Wang, Savvides, and Zhang}]{he2019bounding}
He, Y.; Zhu, C.; Wang, J.; Savvides, M.; and Zhang, X. 2019.
\newblock Bounding box regression with uncertainty for accurate object detection.
\newblock In \emph{Proceedings of the ieee/cvf conference on computer vision and pattern recognition}, 2888--2897.

\bibitem[{Hsu et~al.(2019)Hsu, Hsu, Tsai, Lin, and Chuang}]{hsu2019weakly}
Hsu, C.-C.; Hsu, K.-J.; Tsai, C.-C.; Lin, Y.-Y.; and Chuang, Y.-Y. 2019.
\newblock Weakly supervised instance segmentation using the bounding box tightness prior.
\newblock \emph{Advances in Neural Information Processing Systems}, 32.

\bibitem[{Jha et~al.(2020)Jha, Smedsrud, Riegler, Halvorsen, de~Lange, Johansen, and Johansen}]{jha2020kvasir}
Jha, D.; Smedsrud, P.~H.; Riegler, M.~A.; Halvorsen, P.; de~Lange, T.; Johansen, D.; and Johansen, H.~D. 2020.
\newblock Kvasir-seg: A segmented polyp dataset.
\newblock In \emph{MultiMedia Modeling: 26th International Conference, MMM 2020, Daejeon, South Korea, January 5--8, 2020, Proceedings, Part II 26}, 451--462. Springer.

\bibitem[{Ji et~al.(2023)Ji, Fan, Chou, Dai, Liniger, and Van~Gool}]{ji2023deep}
Ji, G.-P.; Fan, D.-P.; Chou, Y.-C.; Dai, D.; Liniger, A.; and Van~Gool, L. 2023.
\newblock Deep gradient learning for efficient camouflaged object detection.
\newblock \emph{Machine Intelligence Research}, 20(1): 92--108.

\bibitem[{Ji et~al.(2024)Ji, Liu, Xu, Barnes, Khan, Khan, and Fan}]{ji2024frontiers}
Ji, G.-P.; Liu, J.; Xu, P.; Barnes, N.; Khan, F.~S.; Khan, S.; and Fan, D.-P. 2024.
\newblock Frontiers in Intelligent Colonoscopy.
\newblock \emph{arXiv preprint arXiv:2410.17241}.

\bibitem[{Jiang et~al.(2018)Jiang, Zhou, Leung, Li, and Fei-Fei}]{jiang2018mentornet}
Jiang, L.; Zhou, Z.; Leung, T.; Li, L.-J.; and Fei-Fei, L. 2018.
\newblock Mentornet: Learning data-driven curriculum for very deep neural networks on corrupted labels.
\newblock In \emph{International conference on machine learning}, 2304--2313. PMLR.

\bibitem[{Kulharia et~al.(2020)Kulharia, Chandra, Agrawal, Torr, and Tyagi}]{kulharia2020box2seg}
Kulharia, V.; Chandra, S.; Agrawal, A.; Torr, P.; and Tyagi, A. 2020.
\newblock Box2seg: Attention weighted loss and discriminative feature learning for weakly supervised segmentation.
\newblock In \emph{European Conference on Computer Vision}, 290--308. Springer.

\bibitem[{Lan et~al.(2023)Lan, Yang, Yu, Wu, Alvarez, and Anandkumar}]{lan2023vision}
Lan, S.; Yang, X.; Yu, Z.; Wu, Z.; Alvarez, J.~M.; and Anandkumar, A. 2023.
\newblock Vision transformers are good mask auto-labelers.
\newblock In \emph{Proceedings of the IEEE/CVF Conference on Computer Vision and Pattern Recognition}, 23745--23755.

\bibitem[{Lin et~al.(2014)Lin, Maire, Belongie, Hays, Perona, Ramanan, Doll{\'a}r, and Zitnick}]{lin2014microsoft}
Lin, T.-Y.; Maire, M.; Belongie, S.; Hays, J.; Perona, P.; Ramanan, D.; Doll{\'a}r, P.; and Zitnick, C.~L. 2014.
\newblock Microsoft coco: Common objects in context.
\newblock In \emph{Computer Vision--ECCV 2014: 13th European Conference, Zurich, Switzerland, September 6-12, 2014, Proceedings, Part V 13}, 740--755. Springer.

\bibitem[{Loshchilov and Hutter(2017)}]{loshchilov2017decoupled}
Loshchilov, I.; and Hutter, F. 2017.
\newblock Decoupled weight decay regularization.
\newblock \emph{arXiv preprint arXiv:1711.05101}.

\bibitem[{Ma et~al.(2020)Ma, Huang, Wang, Romano, Erfani, and Bailey}]{ma2020normalized}
Ma, X.; Huang, H.; Wang, Y.; Romano, S.; Erfani, S.; and Bailey, J. 2020.
\newblock Normalized loss functions for deep learning with noisy labels.
\newblock In \emph{International conference on machine learning}, 6543--6553. PMLR.

\bibitem[{Ma et~al.(2018)Ma, Wang, Houle, Zhou, Erfani, Xia, Wijewickrema, and Bailey}]{ma2018dimensionality}
Ma, X.; Wang, Y.; Houle, M.~E.; Zhou, S.; Erfani, S.; Xia, S.; Wijewickrema, S.; and Bailey, J. 2018.
\newblock Dimensionality-driven learning with noisy labels.
\newblock In \emph{International Conference on Machine Learning}, 3355--3364. PMLR.

\bibitem[{Mahani et~al.(2022)Mahani, Li, Evangelou, Sotiropolous, Morgan, French, and Chen}]{mahani2022bounding}
Mahani, G.~K.; Li, R.; Evangelou, N.; Sotiropolous, S.; Morgan, P.~S.; French, A.~P.; and Chen, X. 2022.
\newblock Bounding box based weakly supervised deep convolutional neural network for medical image segmentation using an uncertainty guided and spatially constrained loss.
\newblock In \emph{2022 IEEE 19th International Symposium on Biomedical Imaging (ISBI)}, 1--5. IEEE.

\bibitem[{Paszke et~al.(2019)Paszke, Gross, Massa, Lerer, Bradbury, Chanan, Killeen, Lin, Gimelshein, Antiga et~al.}]{paszke2019pytorch}
Paszke, A.; Gross, S.; Massa, F.; Lerer, A.; Bradbury, J.; Chanan, G.; Killeen, T.; Lin, Z.; Gimelshein, N.; Antiga, L.; et~al. 2019.
\newblock Pytorch: An imperative style, high-performance deep learning library.
\newblock \emph{Advances in neural information processing systems}, 32.

\bibitem[{Qadir et~al.(2021)Qadir, Shin, Solhusvik, Bergsland, Aabakken, and Balasingham}]{qadir2021toward}
Qadir, H.~A.; Shin, Y.; Solhusvik, J.; Bergsland, J.; Aabakken, L.; and Balasingham, I. 2021.
\newblock Toward real-time polyp detection using fully CNNs for 2D Gaussian shapes prediction.
\newblock \emph{Medical Image Analysis}, 68: 101897.

\bibitem[{Rother, Kolmogorov, and Blake(2004)}]{rother2004grabcut}
Rother, C.; Kolmogorov, V.; and Blake, A. 2004.
\newblock " GrabCut" interactive foreground extraction using iterated graph cuts.
\newblock \emph{ACM transactions on graphics (TOG)}, 23(3): 309--314.

\bibitem[{Silva et~al.(2014)Silva, Histace, Romain, Dray, and Granado}]{silva2014toward}
Silva, J.; Histace, A.; Romain, O.; Dray, X.; and Granado, B. 2014.
\newblock Toward embedded detection of polyps in wce images for early diagnosis of colorectal cancer.
\newblock \emph{International journal of computer assisted radiology and surgery}, 9: 283--293.

\bibitem[{Song, Kim, and Lee(2019)}]{song2019selfie}
Song, H.; Kim, M.; and Lee, J.-G. 2019.
\newblock Selfie: Refurbishing unclean samples for robust deep learning.
\newblock In \emph{International Conference on Machine Learning}, 5907--5915. PMLR.

\bibitem[{Tajbakhsh, Gurudu, and Liang(2015)}]{tajbakhsh2015automated}
Tajbakhsh, N.; Gurudu, S.~R.; and Liang, J. 2015.
\newblock Automated polyp detection in colonoscopy videos using shape and context information.
\newblock \emph{IEEE transactions on medical imaging}, 35(2): 630--644.

\bibitem[{Tian et~al.(2021)Tian, Shen, Wang, and Chen}]{tian2021boxinst}
Tian, Z.; Shen, C.; Wang, X.; and Chen, H. 2021.
\newblock Boxinst: High-performance instance segmentation with box annotations.
\newblock In \emph{Proceedings of the IEEE/CVF Conference on Computer Vision and Pattern Recognition}, 5443--5452.

\bibitem[{V{\'a}zquez et~al.(2017)V{\'a}zquez, Bernal, S{\'a}nchez, Fern{\'a}ndez-Esparrach, L{\'o}pez, Romero, Drozdzal, Courville et~al.}]{vazquez2017benchmark}
V{\'a}zquez, D.; Bernal, J.; S{\'a}nchez, F.~J.; Fern{\'a}ndez-Esparrach, G.; L{\'o}pez, A.~M.; Romero, A.; Drozdzal, M.; Courville, A.; et~al. 2017.
\newblock A benchmark for endoluminal scene segmentation of colonoscopy images.
\newblock \emph{Journal of healthcare engineering}, 2017.

\bibitem[{Wang and Xia(2021)}]{wang2021bounding}
Wang, J.; and Xia, B. 2021.
\newblock Bounding box tightness prior for weakly supervised image segmentation.
\newblock In \emph{International conference on medical image computing and computer-assisted intervention}, 526--536. Springer.

\bibitem[{Wang and Xia(2022)}]{wang2022polar}
Wang, J.; and Xia, B. 2022.
\newblock Polar transformation based multiple instance learning assisting weakly supervised image segmentation with loose bounding box annotations.
\newblock \emph{arXiv preprint arXiv:2203.06000}.

\bibitem[{Wang et~al.(2019)Wang, Ma, Chen, Luo, Yi, and Bailey}]{wang2019symmetric}
Wang, Y.; Ma, X.; Chen, Z.; Luo, Y.; Yi, J.; and Bailey, J. 2019.
\newblock Symmetric cross entropy for robust learning with noisy labels.
\newblock In \emph{Proceedings of the IEEE/CVF international conference on computer vision}, 322--330.

\bibitem[{Wang et~al.(2023)Wang, Hu, Shi, He, He, Dai, Li, Zhang, Li, Liu et~al.}]{wang2023iboxcla}
Wang, Z.; Hu, Q.; Shi, H.; He, L.; He, M.; Dai, W.; Li, T.; Zhang, Y.; Li, D.; Liu, M.; et~al. 2023.
\newblock IBoxCLA: Towards Robust Box-supervised Segmentation of Polyp via Improved Box-dice and Contrastive Latent-anchors.
\newblock \emph{arXiv preprint arXiv:2310.07248}.

\bibitem[{Wei et~al.(2023)Wei, Hu, Cui, Zhou, and Li}]{wei2023weakpolyp}
Wei, J.; Hu, Y.; Cui, S.; Zhou, S.~K.; and Li, Z. 2023.
\newblock Weakpolyp: You only look bounding box for polyp segmentation.
\newblock In \emph{International Conference on Medical Image Computing and Computer-Assisted Intervention}, 757--766. Springer.

\bibitem[{Wei et~al.(2021)Wei, Hu, Zhang, Li, Zhou, and Cui}]{wei2021shallow}
Wei, J.; Hu, Y.; Zhang, R.; Li, Z.; Zhou, S.~K.; and Cui, S. 2021.
\newblock Shallow attention network for polyp segmentation.
\newblock In \emph{International Conference on Medical Image Computing and Computer-Assisted Intervention}, 699--708. Springer.

\bibitem[{Williams et~al.(2004)Williams, Poulter, Brown, Davis, McInnes, Potter, Sever, and McG~Thom}]{williams2004guidelines}
Williams, B.; Poulter, N.; Brown, M.; Davis, M.; McInnes, G.; Potter, J.; Sever, P.; and McG~Thom, S. 2004.
\newblock Guidelines for management of hypertension: report of the fourth working party of the British Hypertension Society, 2004—BHS IV.
\newblock \emph{Journal of human hypertension}, 18(3): 139--185.

\bibitem[{Wu et~al.(2023)Wu, Chen, Yu, Li, Han, and Jiao}]{wu2023spatial}
Wu, D.; Chen, P.; Yu, X.; Li, G.; Han, Z.; and Jiao, J. 2023.
\newblock Spatial Self-Distillation for Object Detection with Inaccurate Bounding Boxes.
\newblock In \emph{Proceedings of the IEEE/CVF International Conference on Computer Vision}, 6855--6865.

\bibitem[{Xu et~al.(2021)Xu, Zhu, Yang, and Wu}]{xu2021training}
Xu, Y.; Zhu, L.; Yang, Y.; and Wu, F. 2021.
\newblock Training robust object detectors from noisy category labels and imprecise bounding boxes.
\newblock \emph{IEEE Transactions on Image Processing}, 30: 5782--5792.

\bibitem[{Zhang et~al.(2022)Zhang, Fu, Zheng, Zhang, Zhao, and Sham}]{zhang2022hsnet}
Zhang, W.; Fu, C.; Zheng, Y.; Zhang, F.; Zhao, Y.; and Sham, C.-W. 2022.
\newblock HSNet: A hybrid semantic network for polyp segmentation.
\newblock \emph{Computers in biology and medicine}, 150: 106173.

\bibitem[{Zhao, Zhang, and Lu(2021)}]{zhao2021automatic}
Zhao, X.; Zhang, L.; and Lu, H. 2021.
\newblock Automatic polyp segmentation via multi-scale subtraction network.
\newblock In \emph{Medical Image Computing and Computer Assisted Intervention--MICCAI 2021: 24th International Conference, Strasbourg, France, September 27--October 1, 2021, Proceedings, Part I 24}, 120--130. Springer.

\bibitem[{Zhu et~al.(2023)Zhu, Shi, Zhao, Wang, Qiao, and An}]{zhu2023contrast}
Zhu, Z.; Shi, J.; Zhao, M.; Wang, Z.; Qiao, L.; and An, H. 2023.
\newblock Contrast Learning Based Robust Framework for Weakly Supervised Medical Image Segmentation with Coarse Bounding Box Annotations.
\newblock In \emph{International Workshop on Computational Mathematics Modeling in Cancer Analysis}, 110--119. Springer.

\end{thebibliography}

\end{document}